# Face Recognition in Unconstrained Conditions: A Systematic Review

ANDREW JASON SHEPLEY, Charles Darwin University, Australia

ABSTRACT
Face recognition is a biometric which is attracting significant research, commercial and government interest, as it provides a discreet, non-intrusive way of detecting, and recognizing individuals, without need for the subject's knowledge or consent. This is due to reduced cost, and evolution in hardware and algorithms which have improved their ability to handle unconstrained conditions. Evidently affordable and efficient applications are required. However, there is much debate over which methods are most appropriate, particularly in the context of the growing importance of deep neural network-based face recognition systems. This systematic review attempts to provide clarity on both issues by organizing the plethora of research and data in this field to clarify current research trends, state-of-the-art methods, and provides an outline of their benefits and shortcomings. Overall, this research covered 1,330 relevant studies, showing an increase of over 200% in research interest in the field of face recognition over the past 6 years. Our results also demonstrated that deep learning methods are the prime focus of modern research due to improvements in hardware databases and increasing understanding of neural networks. In contrast, traditional methods have lost favor amongst researchers due to their inherent limitations in accuracy, and lack of efficiency
when handling large amounts of data.

## 1    INTRODUCTION

The development of accurate and efficient face recognition systems for use in unconstrained conditions is an area of high research interest. This is largely due to the desire of governments, business and consumers to improve efficiency of a wide range of systems used in everyday life, industry and security [38]. Because of this interest, and the recent developments in hardware such as graphical processing units (GPUs), a large number of publications and academic contributions have been made over the past few years, as illustrated by Figure 1. This data is largely unorganized and confusion over the most appropriate and effective face recognition systems persists [46]. Essentially, the goal of face recognition is to identify whether there is a face in a given image, and then identify
who the face belongs to, irrespective of environmental conditions, distance from camera, weather, make-up and other more complex factors such as age [37]. These tasks are currently performed at substandard accuracy in tasks such as airport security [24], gimmicks like SnapChat [59] and practical applications such as user authentication [42]. However, despite often poor performance in real life conditions [1], face recognition has been touted as the biometric of the future due to its non-invasive and discreet nature [15].

---

Author's address: Andrew Jason Shepley, Charles Darwin University, Ellengowan Drive, Casuarina, NT, 0810, Australia, ashepley@myune.edu.au.

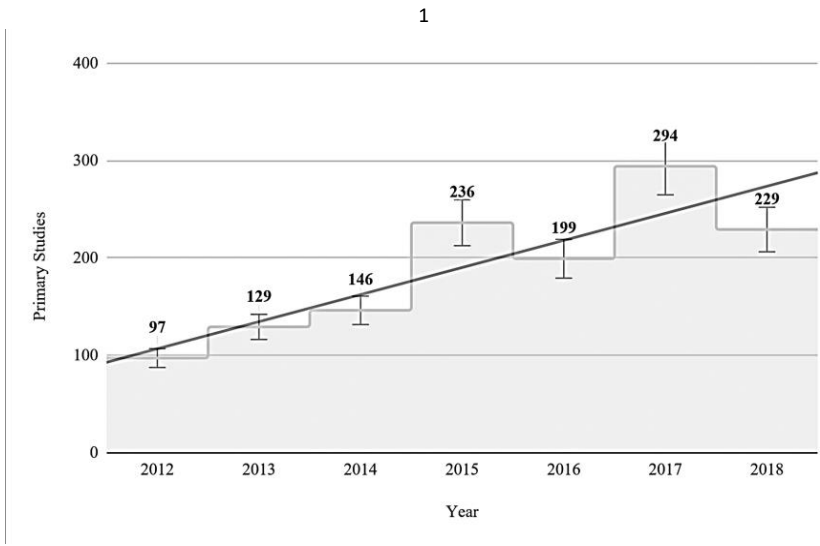

Fig. 1. Increasing number of publications in the field of face recognition.

Due to the significant interest in this field of research, the volume in research studies has risen by 203% over the 2012-2017 period, from 97 published studies in 2012 to 294 in 2017. However this significant volume of research data is not organized in a way which provides a clear indication of current trends or direction on future research and prospects. Most literature reviews which were examined do not follow a systematic approach [38], which renders them susceptible to bias [27] and do not specifically address unconstrained face recognition. Furthermore, as a whole, it is our opinion that they provide no clear indication of which face recognition technologies are most appropriate in real life applications. Thus, this review is necessary to show not only which face recognition methods are preferred by modern researchers, but to also provide an insight into why movements in research direction have happened, based on the strengths and limitations of the many variations in face recognition techniques, the current state-of-the-art technologies and the challenges faced by developers of face recognition solutions in unconstrained conditions.

Therefore the goal of this article is to report the findings of a systematic literature review which will attempt to clarify these issues. The article will be structured as follows: Section 2 will provide definitions and an overview of the main concepts referred to in this review. Section 3 will describe the objectives of the review, list the research questions, explain the search strategy and the selection process. Section 4 will outline the evaluation criteria and data extraction strategy. Section 5 will present results while Section 6 provides a conclusive summary of the findings of the review, emphasizing directions on future research work.

## 2 REVIEW PROTOCOL AND BACKGROUND CONCEPTS

## 2.1 Systematic Review Protocol

A systematic review is a review which identifies, synthesizes and critically appraises all relevant research in the scope of a set of clearly formulated research questions, using a systematic and reproducible method. Studies and works which are determined to be relevant research are referred to as primary studies. This review is based on the guidelines specifically developed by [27] for systematic reviews in the field of software engineering. The most important stage in a systematic review identified by [27] is the determination of the main research questions, followed by the identification of keywords to be used in defining the search string. We identified keywords by reference to studies of interest, and synonyms of the keywords used. A search string, referred to as an automatic search is then constructed. Complementary search methods such as conducting a manual search is used to further improve collection of primary studies.

However, both automatic and manual searches are limited by the arbitrary selection of databases and other research forums, the effectiveness of the database interfaces, and the nature of the search string used [27]. Thus, a secondary strategy referred to as snowballing was also employed. Snowballing involves collecting a set of primary papers, then referring to the reference lists in these papers to identify further primary papers [60]. This process is repeated until all primary papers are found. It is an effective means of collecting all primary studies relevant to the research questions.

Thus, this systematic review will use an automatic and manual search to identify primary studies, followed by the implementation of an effective snowballing search strategy to further refine the set of primary studies to effectively answer the research questions.

## 2.2 General Definitions

Face recognition can be defined as a category of biometric software which functions to identify or verify the identity of one or more persons in an image [38]. It is a generic term used to describe a wide range of technologies including still image, video-based [49], 3D [4] and infrared [31] recognition techniques in both constrained and unconstrained conditions. Figure 2 shows the general face recognition pipeline used in most unconstrained face recognition systems.

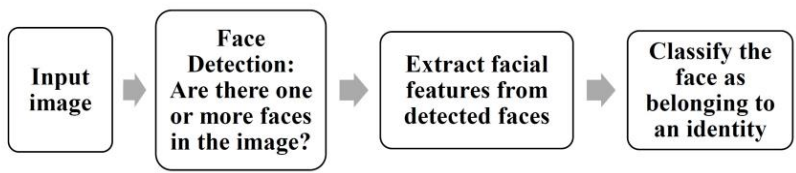

Fig. 2. General Face Recognition Pipeline.

*2.2.1 Unconstrained Conditions:* Lack of control over environmental factors such as lighting, pose, occlusion and distance from the camera [7]. The subject of this research is real-time face recognition in unconstrained conditions, which is relevant to real-life applications in a wide range of contexts including security, consumer applications, and user verification.

## 3  SYSTEMATIC REVIEW METHODOLOGY

The methodology used to conduct this systematic review is described below, and summarized in Figure 3

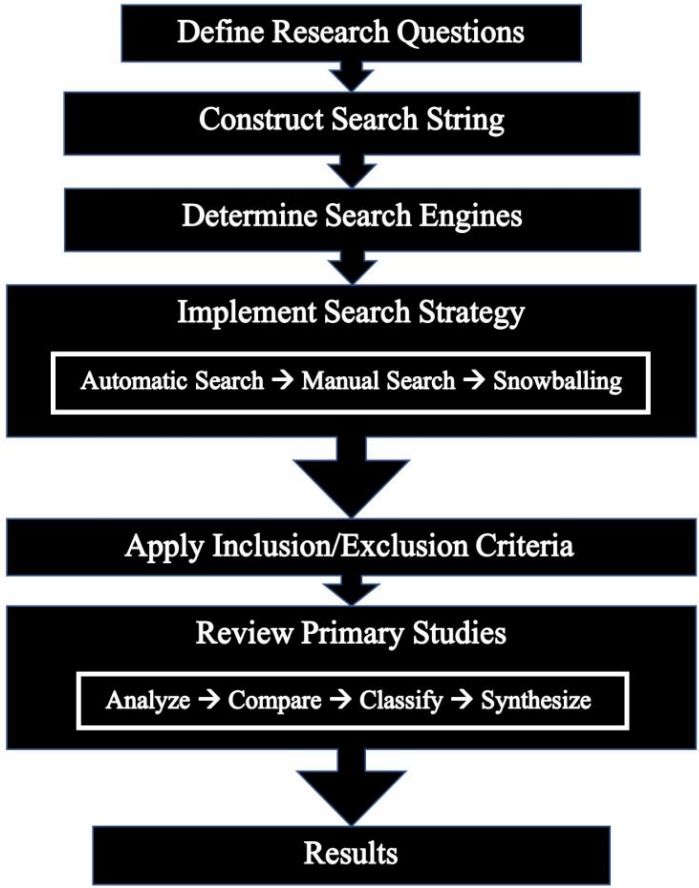

Fig. 3. Systematic review methodology.

### 3.1  Research Questions

We formulated a set of research questions based on a background overview and understanding of the current uses and applications of face recognition. These questions will aim to address the purpose of this systematic review, namely, to provide a clear picture of the current trends in research, identify and evaluate state-of-the-art methods in unconstrained face recognition, and the challenges faced by modern researchers:

(1) *Which methods are most commonly used to achieve unconstrained face recognition and which obtain state of the art results?*
(2) *What are the challenges currently facing unconstrained face recognition and where could it be used in the future?*
(3) *Which datasets are currently used to develop and test face recognition techniques?*
(4) *What are the most commonly used tools for unconstrained face recognition?*

### 3.2 Defining the Search String

During background research in the area of interest, the following set of keywords, including their synonyms, were identified as relating directly to unconstrained face recognition:

(1) Face OR Facial AND
(2) Recognition OR Identification OR Verification AND
(3) Methodologies OR Techniques OR Systems OR Processes AND
(4) Unconstrained OR 'in the wild'

We used these keywords in both the automatic and manual search phases, only targeting studies published between 2012 and 2018 to make sure the review produced results which clearly show trends and challenges in modern research.

### 3.3 Defining the Search Engines

In order to provide a comprehensive coverage of all relevant research, the following academic databases were automatically and manually searched using the search string:

(1) IEEEXplore(www.ieeexplore.com.br)
(2) ScienceDirect(www.sciencedirect.com)
(3) SpringerLink(www.link-springer-com)
(4) ACM(www.portal.acm.org)
(5) Scopus(www.scopus.com)
(6) WileyOnlineLibrary(www.onlinelibrary.wiley.com)

These databases are online and limited to studies published in English.

### 3.4 Conducting the Search

We then used the search string to execute the search on the chosen databases. Both an automatic search using the search string, and a manual search were conducted to reveal primary papers which may have been missed. This was followed by snowballing, which we are confident revealed the entire set of primary studies on unconstrained face recognition. This included all relevant journals, conference proceedings, technical papers and other relevant literature. The research was conducted between February and September 2018, and all results were downloaded in RIS format for easy reference management in Mendeley, such as removal of duplicate studies. The unrefined search results have been provided in Table 1.

Table 1. Unrefined Search Results

| DATABASE | NUMBER OF STUDIES |
|---|---|
| IEEE Xplore | 1,117 |
| ScienceDirect | 1,367 |
| Springer Link | 893 |
| ACM | 457 |
| Scopus | 503 |
| Wiley Online | 237 |
| TOTAL | 4574 |

### 3.5 Inclusion and Exclusion Criteria

This stage involved the application of inclusion and exclusion criteria used to determine which studies would proceed to the next stage of review. We aligned the criteria with the research questions, in order to achieve the aim of the systematic review. The criteria are presented in Table 2. Due to the large volume of potentially relevant studies, the first phase involved elimination of studies based only on the titles, where studies clearly did not fit into the scope of the review. In the second phase, the title and abstract of each remaining study was used to determine compliance with the inclusion or exclusion criteria. All papers that were deemed doubtful were downloaded and reviewed by several team members to ensure compliance with the inclusion and exclusion criteria. The studies which passed all stages of refinement were then downloaded, and the introduction and conclusion were reviewed according to the same criteria. All papers which passed this final stage were then read in their entirety and included or excluded according to the same criteria.

Table 2. Inclusion and Exclusion Criteria

| INCLUSION | EXCLUSION |
|---|---|
| All primary studies in English language | Exclude non-English language studies |
| All studies which are at least 25 percent different to any other study | Exclude duplicate studies |
| Only include studies that directly address unconstrained face recognition | Exclude studies that do not relate to face recognition or relate only to constrained face recognition |
| Include all peer-reviewed journals and conference papers, book chapters and technical papers | Exclude very short papers, posters, presentations, and editorials |
| Include all relevant methodology and technical studies | Exclude secondary studies such as surveys and reviews |
| Include all literature that answers one or more research questions | Literature that does not answer at least one research question |
| Include all studies published between 2012 and 2018 | Exclude all studies published prior to 2008 |

### 3.6 Refined Primary Studies List

Subsequent to the execution of the search string, and application of the inclusion and exclusion criteria, the set of refined primary studies indicated in Table 3 were obtained from the initial search results. In total, 1330 primary studies were identified, and all papers were used for data extraction and evaluation according to the research questions outlined in Section 2.

## 4 DATA EVALUTION AND EXTRACTION

The evaluation of the final set of primary studies was based on the following data types extracted from each primary study:

(1) Title
(2) Year
(3) Face detection technique
(4) Face identification/verification technique
(5) Challenge addressed [e.g occlusion, pose, lighting, etc.]
(6) Tools used [e.g. programming language, network, hardware, etc.]
(7) Database used
(8) Accuracy for detection, verification and identification

Table 3. Refined Search Results

| DATABASE | NUMBER OF STUDIES |
|---|---|
| IEEE Xplore | 303 |
| ScienceDirect | 376 |
| Springer Link | 267 |
| ACM | 158 |
| Scopus | 153 |
| Wiley Online | 73 |
| TOTAL | 1330 |

This data was extracted from each primary study and used to answer each research question. The results and discussion presented in response to each question will provide a clear evaluation of current research trends in unconstrained face recognition, state-of-the-art methods, and most commonly faced challenges, and will indicate which tools and databases are most commonly employed by researchers. We hope this will address our aim of facilitating further research and development in face recognition technologies.

## 5 RESULTS AND DISCUSSION

This review has revealed a consistent increase in research interest in unconstrained face recognition within the period 2012-2018, as shown by Figure 1. This is due to an increase in usage of face recognition as a biometric in a wide range of applications, the significant improvements in technology such as powerful Graphical Processing Units (GPUs), and the creation of huge annotated face datasets. We anticipate that this interest will continue to increase into the future, thus the need for this review. More specifically, this section aims to provide a comprehensive response to each research question.

### 5.1 Which methods are most commonly used to achieve unconstrained face recognition and which obtain state of the art results?

To succinctly respond to this question, the primary studies have been classified according to the techniques used for unconstrained face recognition and presented in Table 4.

Table 4. Face Recognition Techniques

| TECHNIQUE | NUMBER OF STUDIES |
|---|---|
| Traditional handcrafted features combined with machine learning | 153 |
| Learned Features & Deep Convolutional Neural Networks | 797 |
| Combination of deep neural networks and local handcrafted features | 338 |
| Infrared and Thermal | 12 |

Table 4 identifies the general categories of methods used to achieve unconstrained face recognition. These methods are used to attempt to distinguish between objects classified as human faces and other objects in the background. They also attempt to identify who the face belongs to, or verify that a face belongs to a particular individual. They rely on a range of structural and texture-based feature representation techniques, including traditional handcrafted features such as Haar features [55] and Local Binary Patterns [3], fiducial points [36], and learned features [54]. Classification of faces is usually achieved using machine learning algorithms, or more recently, neural networks [44].

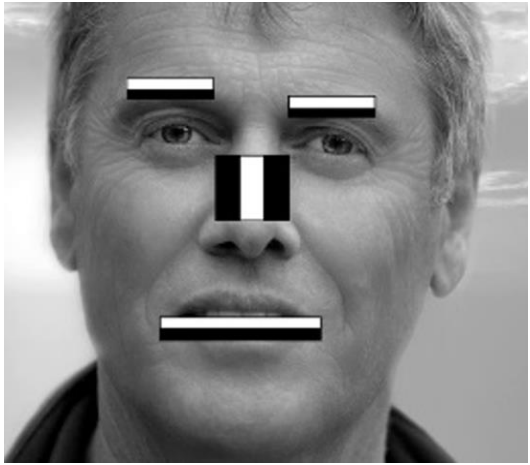

Fig. 4. An example of traditional handcrafted Haar features [20].

*5.1.1 Traditional or local handcrafted features:* Chosen features which represent properties of the image, such as edges and corners, which are detected using algorithms. A good example of this is the use of Haar features in face detection by Viola Jones [55] as shown in Figure 4. Usually, machine learning algorithms such as Support Vector Machines (SVM) are then used to classify a collection of features as comprising a face, or not [18].

*5.1.2 Learned Features:* Automatically emerge from data, and evolve via a complex model such as a neural network [54]. Although parameters are set to control the learning process, there is a paucity of information regarding the actual features that are learned, as they continue to evolve for as long as they are optimized using a process such as gradient descent [44]. Learned features are very useful in face recognition, due to the complexity and non-linearity of human features, which can be learned using large databases of images of people in a wide range of poses, and expressions [52]. Learned features using DCNNs is the preferred means of achieving face recognition, with 797 out of 1,330 or 61% of primary studies focusing on this methodology, and an additional 338, or 26% combining DCNNs with traditional features to achieve superior results to traditional systems alone. This data correlates with the improvements in accuracy and reliability of deep learning based methods as availability of large databases increases [35], in contrast with the decline in performance of traditional systems, which become inefficient when processes large amounts of data [35], as illustrated by Figure 5.

Although accurate results can be obtained using traditional handcrafted methods [3, 17], especially under constrained conditions, this data suggests that deep convolutional neural networks have become the most widely used technique in unconstrained face recognition, due to their powerful ability to learn distinguishing features [11, 16, 25, 54]. In fact, they have become so effective that they now surpass human ability in face verification [54], and outperform all traditional handcrafted methods [45, 54]. However, DCNNs rely on the use of large databases [26, 61] for training and inference, which makes them computationally expensive [8, 53]. Additionally, training on huge quantities of images takes weeks or months for a large system. Because of this, they can be quite impractical to use in real life situations, which has triggered the creation of traditional and deep learning hybrids which [6, 33] to provide solutions that are both efficient and accurate, while other methods based on traditional handcrafted features are still used in some cases [3].

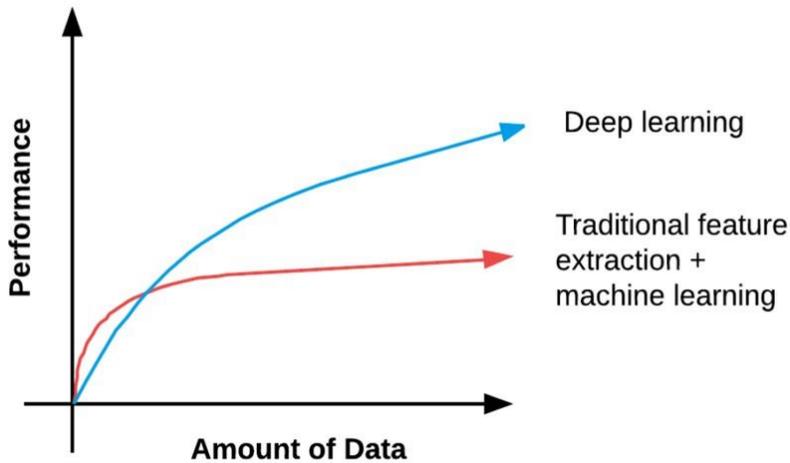

Fig. 5. A comparison of the performance of deep learning based face recognition with traditional feature based methods as the availability of large databases increases. [43].

Following is an outline and evaluation the state of the art methods used for face detection, and face identification/verification. Table 5 shows the top 5 state-of-the-art in face detection methods, while Table 6 shows the top 4 face identification and verification methods. It is important to note that all 10 methods are based on deep convolutional neural networks.

Table 5. Top 5 Face Detection Methods

| METHOD | WIDERFACE |
| --- | --- |
| FAN[58] | 0.885 |
| S3FD[64] | 0.858 |
| SSH[34] | 0.844 |
| HR[22] | 0.819 |
| ScaleFaces[62] | 0.764 |

*5.1.3 State-of-the-Art Face Detection - Face Attention Network (FAN) [58]:* proposed a novel face detector focused on improving detection of faces degraded by occlusion, e.g wearing sunglasses or a face mask, without increasing the rate of false positives. FAN is a one shot detector based on the simple, dense object detector, RetinaNet [30]. It is based on ResNet-50, and uses a feature pyramid network, and several layers to address faces of varying scales, as shown in Figure 6. Significantly, it proposes an anchor-level attention model, optimized using a multi-task loss function, which highlights features within facial regions in different layers to reduce instances of false positives. Attention models attempt to mimic the human brain by focusing computational resources on specific regions at high resolution while appreciating the surrounding regions at low resolution, rather than systematically exploring the entire visual field. This resulted in the achievement of stateof-the-art results in face detection, and is the first time that a single-shot detector has outperformed more conventional two-stage detectors popularized by the success of the Faster R-CNN [40].

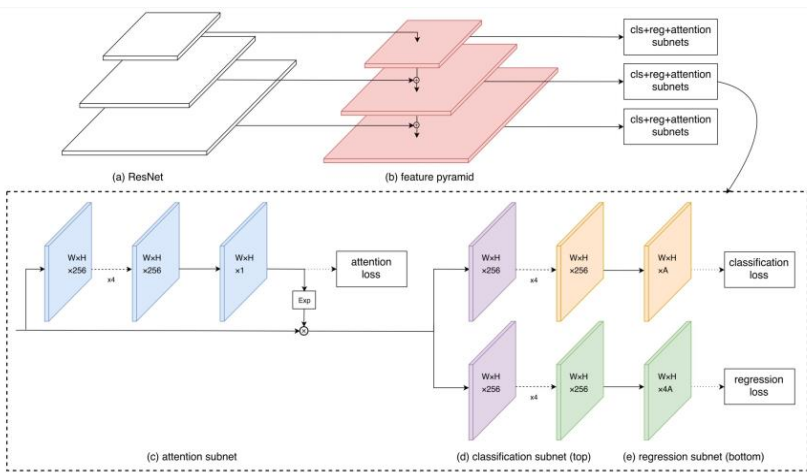

Fig. 6. The FAN Network Architecture [58].

Random-crop is further used to augment the WiderFace dataset, resulting in an increase in the number of occluded faces in the dataset. This was used to train the network to be robust to occlusion. FAN was evaluated on the WiderFace dataset, outperforming other state of the art methods such as [34, 64] and [62]. Examples of its effectiveness on this dataset are illustrated by Figure 7. Recently, the Faster R-CNN based Enhanced Region Proposal Network, proposed by [12] has achieved state-of-the-art results in object detection. Possible future research may include adapting this network specifically for face detection.

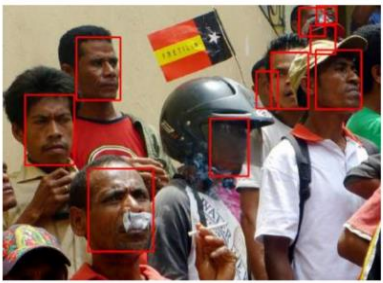
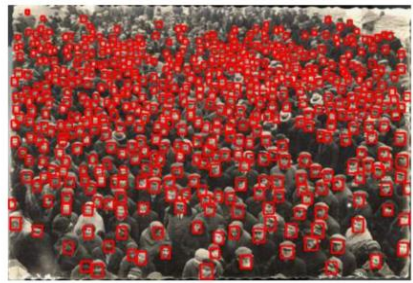

Fig. 7. Examples of the experimental results of FAN on the WiderFace dataset [58].

*5.1.4 State-of-the-art Face Identification and Verification - ArcFace [13]:* One of the most important components of a DCNN-based face recognition pipeline is the loss function. The loss function calculates the prediction error, allowing the DCNN to be trained using an error minimization process such as gradient descent. Its goal is to minimize intraclass variation (same identities) and maximize interclass variation (different identities). ArcFace essentially contributed a novel loss function which uses an Additive Angular Margin Loss to obtain highly discriminative facial features. It was evaluated on many datasets and benchmarks, including WiderFace, clearly outperforming all previous state-of-the-art methods [14, 45, 51, 57] as shown in Table 6.

Table 6. Top 5 Face Identification and Verification Methods

| METHOD | Verification | Identification |
| --- | --- | --- |
| ArcFace (LResNet100E-IR) [13] | 0.998 (LFW) 0.985 (MF1) | 0.833 (MF1) |
| CosFace [57] | 0.997 (LFW) 0.980 (MF1) | 0.845 (LFW) |
| FaceNet [45] | 0.996 (LFW) 0.865 (MF1) | 0.705 (LFW) |
| DCFL [14] | 0.996 | - |
| DeepID2+ [51] | 0.995 | - |

The use of the ArcFace loss function in DCNN-based face recognition offers various benefits. Its creators claim it is engaging, easy to implement, and effective. It improves feature representation by optimizing the geodesic (shortest possible distance between two points on a curved surface) distance margin to reflect the direct relationship between the angle and corresponding arc [13]. It consequently achieves state-of-the art accuracy and precision, at an acceptable computational speed using standard GPUs. Furthermore, it is easy to implement in standard deep learning frameworks as shown by Figure 8. It offers greater robustness than the traditionally popular Softmax loss function, which fails to optimise feature embeddings resulting in inadequate similarity between intraclass samples, and suboptimal difference between interclass variations. Thus, ArcFace rectifies the performance gap in unconstrained conditions [13].

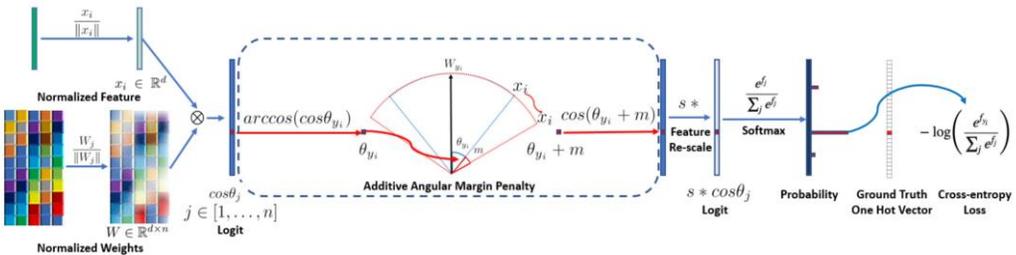

Fig. 8. Implementation of the ArcFace Loss Function into a DCNN for training [13]

## 5.2      What are the challenges currently facing unconstrained face recognition?

Unconstrained face recognition systems are broadly applied in government, industry, commercial and consumer security solutions. They are particularly useful in law enforcement applications, and have also been employed by social media corporations, user feature enhancement, and in consumer applications such as cameras and phones. Very recently, high accuracy and reliability of systems has meant companies have been using face recognition instead of ID cards to enable employees to access premises and systems [41]. Future uses could encompass prevention of theft, employee performance monitoring, use of face recognition as a digital signature, and many other application.

Clearly, identification and verification tasks in these contexts requires face recognition systems to effectively handle situations involving both environmental occlusion [9, 56], variations in pose [32, 50] and expression [29] and deliberate attempts to circumvent security systems [39]. They also need to handle natural changes in appearance caused by weight loss or gain, aging, and other physical changes [10, 65]. This poses serious problems for developers of unconstrained face recognition systems due to the very large range in variables. This review classified primary studies based on which particular difficulties they attempted to address as shown in Figure 9. These results indicate that the majority of research attempts to handle unconstrained environmental issues generally. More specifically, 39% of studies either did not specifically mention the effects of unconstrained environmental effects, or addressed all conditions generally. This was particularly true for studies which use deep learning based methodologies. We infer that this is because unconstrained features are learned during training, and thus do not need to be directly addressed. This provides for more robustness to a broader range of environmental factors [45]. In contrast, the majority of studies which addressed a particular environmental issue involved the use of traditional features. This reflects the nature of hand-crafted features: they attempt to impose arbitrarily chosen patterns onto data rather than allowing natural, unconstrained features to emerge from the data. Thus, in our opinion, a relationship between lack of robustness of traditional methods and the decline in research interest in these methods can be established.

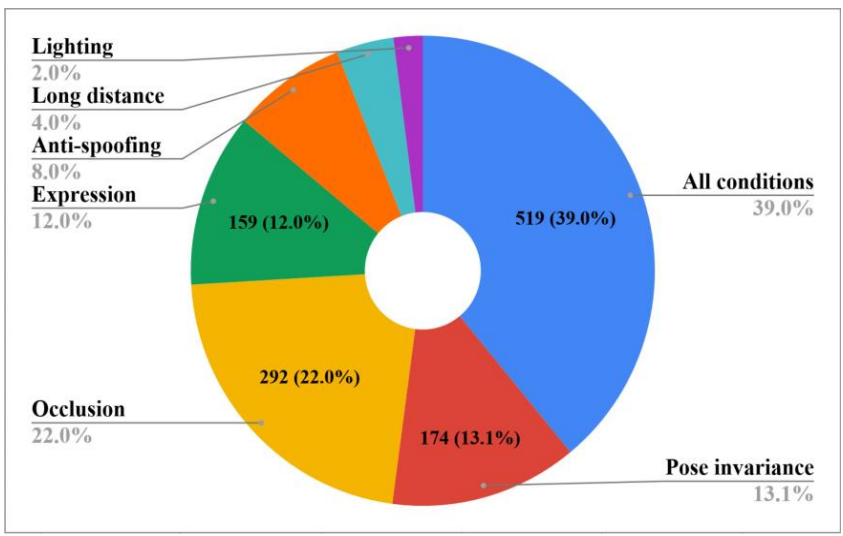

Fig. 9. Environmental challenge most commonly addressed by research.

The problem of deliberate or malicious circumvention of face recognition systems to access confidential data or restricted areas is of particular importance. Techniques used for circumvention of face recognition range from anti-face recognition masks, use of lighting to inhibit the cameras from capturing quality images, and the use of patterns to confuse face recognition systems. This challenge has attracted specific research into the areas of anti-fraud and anti-spoofing face recognition systems, which accounts for 106 out of 1330 studies, or 8% of all research. Although some studies claim to be largely successful, especially in detecting the difference between a three-dimensional human face, and a photograph of a human face [39], the data collected as part of this systematic review has highlighted a shortage of algorithms and solutions to this problem. In the context of increasing usage of face recognition as a biometric for security purposes, improving the reliability of face recognition systems is of paramount importance. Thus, one area of future research which should be considered is the development of intelligent face recognition systems which learn to recognize when a system may be compromised, and adapt to threats as they evolve.

Furthermore, a significant proportion of the primary studies indicated that unconstrained 3D face recognition systems would have a distinct advantage over currently used 2D face recognition systems, due to the additional features this would provide [4, 45, 54]. However, the technology used to collect 3D images is prohibitively expensive, and the available databases are very limited and small [4]. Therefore, as the application of face recognition technologies is extended, this shortage will negatively impact on reliability, and accuracy of the techniques used. Rectifying this issue in the future is necessary to prevent sensitive systems from being compromised, as society moves into a more biometric oriented mechanism of managing security and data.

5.3     Which datasets are used to develop and evaluate face recognition techniques?

Large datasets of images containing faces are necessary to the development of effective unconstrained face recognition systems. They are essential to the training of large, DCNNs, which rely on thousands to millions of images to accurately learn distinguishing features [26, 54, 61]. They are also indispensable when testing face recognition systems, to demonstrate a valid comparison of currently used techniques, and proposed improvements [26]. Thus, it is necessary for this review to consider the primary studies in light of the datasets used for development and testing of face recognition systems. Furthermore, it is of high importance that the datasets used contain unconstrained features, such as large variations in pose, expression and illumination, as well as occlusion and scale [26]. Other aspects which need to be considered are the inclusion of images of people from a wide range of ages, races and gender [45].

Table 7. Face Recognition Datasets

| DATASET | WEBSITE | FEATURES |
|---|---|---|
| MegaFace | http://megaface.cs.washington.edu/index.html | 4,700,000 images 672,000 identities |
| WIDER FACE | http://mmlab.ie.cuhk.edu.hk/projects/WIDERFace/ | 32,203 images containing 393,703 faces |

| Labelled Faces in the Wild (LFW) | http://vis-www.cs.umass.edu/lfw/ | 13,233 images 5749 identities |
|---|---|---|
| VGG2-Face | http://www.robots.ox.ac.uk/~vgg/data/vgg_face2/ | 3,310,000 images 9131 identities |
| MS-Celeb-1M | https://www.msceleb.org/ | 10,000,000 images 100,000 identities |
| FaceNet | Private | 100M-200M images, 8,000,000 identities |

Table 7 contains the features of the six most commonly used datasets, which are often accompanied by benchmarking standards. Due to the data-driven nature of the Internet and media collection, these databases are becoming increasingly large, as exemplified by the MegaFace Challenge Dataset and Benchmark [26], which contains 4.7 million images containing 672,000 identities and the MS-Celeb-1M dataset which contains over 10 million images of 100,000 identities. The most commonly used datasets are bundled with benchmarking standards and a testing protocol, and are oriented towards specific types of face recognition tasks, for example the WIDER Face dataset and benchmark [61] is aimed at face detection development, so it contains a large testing set annotated with ground truth bounding boxes. Other databases such as the Labeled Faces in the Wild dataset and benchmark [63] and the VGG-2 dataset are aimed at testing and comparing face verification. Each of these datasets include thousands or millions of images used to train, test and compare face recognition systems. They are a causative factor in the upward in the interest in and use of DCNN based face recognition systems by modern researchers.

These datasets are essential in validating unconstrained face recognition systems, and measuring their ability to handle a wide range of real-life situations. Although some datasets which have contributed to some of the biggest developments in face recognition, such as the FaceNet dataset [45] are private, the majority of these datasets are publicly available or available for research purposes, which facilitates collaboration on the development of effective solutions. Many smaller, more specific face databases have also been developed to handle particular aspects of unconstrained face recognition. These focus on cross-age face recognition [66], long distance face recognition [23], video-based face recognition [48], and those dealing with severe occlusion [61]. However, these are often too small to effectively train a DCNN, although they are useful in evaluating the effectiveness of unconstrained face recognition systems, particularly in specific situations affected by environment degradation.

5.4    What are the most commonly used tools for unconstrained face recognition?

The primary studies analyzed indicate that researchers primarily use Python as the programming language of choice in developing face recognition solutions. Python is a high-level programming language which can be run on multiple platforms including Windows, UNIX, and Mac OS, which is the reason why it is so commonly used. Python also leverages off powerful libraries such as Keras [5] and Tensorflow [2], which contain mathematical and optimization functions useful for modifying and improving existing techniques. Python is not only used for implementation of algorithms, but is also used to evaluate the effectiveness of the unconstrained face recognition systems after development and training, using data science techniques. Researchers focusing on DCNN based face recognition methods rely largely on proven convolutional neural network architectures such as ResNet [19], MobileNet [21], AlexNet [28] and VGGNet [47], which are usually programmed using C, and supplemented using Python.

This systematic review overcomes the shortage in current secondary reviews and surveys by providing an extensive and comprehensive analysis and evaluation of current trends and direction in unconstrained face recognition systems. It encompasses all the techniques currently used, using data to show which methods, tools and databases are preferred by researchers, and what challenges they face. Furthermore, it provides insight into the possible reasons explaining the current trends, and suggests possible research direction based on critical data analysis. It provides further value by highlighting the difficulties and issues facing modern unconstrained face recognition in the context of their expanding application in real life.

Thus, this review provides direction for future research, by providing a clear perspective on the current unconstrained face recognition research field, and the advantages and disadvantages of currently used techniques. It has leveraged the power of six extensive databases to ensure comprehensive coverage of all available techniques, tools and databases. This will provide a basis for future improvement of systems, further research and improvement in unconstrained face recognition. It highlights open issues for future research and investigation, in light of the increasing importance of face recognition in computer vision and the expanding application of unconstrained face recognition in security, consumer and commercial applications and industry.

## 6  CONCLUSIONS AND FUTURE WORK

This systematic review provides a clear perspective on the current state of unconstrained face recognition research. It shows a clear increase in research interest in this field over the past six years, and illustrates using data, the increasing interest in data-driven deep convolutional neural networks in preference of traditional hand-crafted features and machine learning algorithms. It provides a clear picture of the challenges facing modern face recognition and indicates how these challenges are being addressed with the aim of highlighting areas of future interest, and clarifying the current status quo.

## ACKNOWLEDGMENTS

This research is supported by an Australian Government Research Training Program (RTP) Scholarship.

## REFERENCES

[1] 2018. Face firms fight ÃćâĆňĚIJ98claims. *Biometric Technology Today* 2018, 6 (2018), 11–12.
[2] Martín Abadi, Ashish Agarwal, Paul Barham, Eugene Brevdo, Zhifeng Chen, Craig Citro, Gregory S. Corrado, Andy Davis, Jeffrey Dean, Matthieu Devin, Sanjay Ghemawat, Ian J. Goodfellow, Andrew Harp, Geoffrey Irving, Michael Isard, Yangqing Jia, Rafal Józefowicz, Lukasz Kaiser, Manjunath Kudlur, Josh Levenberg, Dan Mané, Rajat Monga, Sherry Moore, Derek Gordon Murray, Chris Olah, Mike Schuster, Jonathon Shlens, Benoit Steiner, Ilya Sutskever, Kunal Talwar, Paul A. Tucker, Vincent Vanhoucke, Vijay Vasudevan, Fernanda B. Viégas, Oriol Vinyals, Pete Warden, Martin Wattenberg, Martin Wicke, Yuan Yu, and Xiaoqiang Zheng. 2016. TensorFlow: Large-Scale Machine Learning on Heterogeneous Distributed Systems. *CoRR* abs/1603.04467 (2016). arXiv:1603.04467 http://arxiv.org/abs/1603.04467

[3] T. Ahonen, A. Hadid, and M. Pietikainen. 2006. Face Description with Local Binary Patterns: Application to Face Recognition. *IEEE Transactions on Pattern Analysis and Machine Intelligence* 28, 12 (Dec 2006), 2037–2041. https://doi.org/10.1109/TPAMI.2006.244

[4] Zhanfu An, Weihong Deng, Tongtong Yuan, and Jiani Hu. 2018. Deep Transfer Network with 3D Morphable Models for Face Recognition. (05 2018), 416–422. https://doi.org/10.1109/FG.2018.00067

[5] Dustin Anderson, Jean-Roch Vlimant, and Maria Spiropulu. 2017. An MPI-Based Python Framework for Distributed Training with Keras. *CoRR* abs/1712.05878 (2017). arXiv:1712.05878 http://arxiv.org/abs/1712.05878

[6] Olasimbo Ayodeji Arigbabu, Saif Mahmood, Sharifah Mumtazah Syed Ahmad, and Abayomi A. Arigbabu. 2016. Smile detection using hybrid face representation. *Journal of Ambient Intelligence and Humanized Computing* 7, 3 (01 Jun 2016), 415–426. https://doi.org/10.1007/s12652-015-0333-4

[7] Lacey Best-Rowden, Hu Han, Charles Otto, Brendan F. Klare, and Anil K. Jain. 2014. Unconstrained Face Recognition: Identifying a Person of Interest From a Media Collection. *Trans. Info. For. Sec.* 9, 12 (Dec. 2014), 2144–2157. https://doi.org/10.1109/TIFS.2014.2359577

[8] Z. BukovÄŋikovÃ¡, D. Sopiak, M. Oravec, and J. PavloviÄŋovÃ¡. 2017. Face verification using convolutional neural networks with Siamese architecture. In *2017 International Symposium ELMAR*. 205–208. https://doi.org/10.23919/ELMAR.2017.8124469

[9] X. P. Burgos-Artizzu, P. Perona, and P. DollÃąr. 2013. Robust Face Landmark Estimation under Occlusion. In *2013 IEEE International Conference on Computer Vision*. 1513–1520. https://doi.org/10.1109/ICCV.2013.191

[10] B. Chen, C. Chen, and W. H. Hsu. 2015. Face Recognition and Retrieval Using Cross-Age Reference Coding With Cross-Age Celebrity Dataset. *IEEE Transactions on Multimedia* 17, 6 (June 2015), 804–815. https://doi.org/10.1109/TMM.2015.2420374

[11] Jun-Cheng Chen, Rajeev Ranjan, Swami Sankaranarayanan, Amit Kumar, Ching-Hui Chen, Vishal M. Patel, Carlos D. Castillo, and Rama Chellappa. 2018. Unconstrained Still/Video-Based Face Verification with Deep Convolutional Neural Networks. *International Journal of Computer Vision* 126, 2 (01 Apr 2018), 272–291. https://doi.org/10.1007/s11263-017-1029-3

[12] Yu Peng Chen, Ying Li, and Gang Wang. 2018. An Enhanced Region Proposal Network for object detection using deep learning method. *PLOS ONE* 13, 9 (09 2018), 1–26. https://doi.org/10.1371/journal.pone.0203897

[13] Jiankang Deng, Jia Guo, and Stefanos Zafeiriou. 2018. ArcFace: Additive Angular Margin Loss for Deep Face Recognition. *CoRR* abs/1801.07698 (2018). arXiv:1801.07698 http://arxiv.org/abs/1801.07698

[14] W. Deng, B. Chen, Y. Fang, and J. Hu. 2017. Deep Correlation Feature Learning for Face Verification in the Wild. *IEEE Signal Processing Letters* 24, 12 (Dec 2017), 1877–1881. https://doi.org/10.1109/LSP.2017.2726105

[15] Alexandra Frean US Business Editor. 2015. Intel sees its future in face recognition and driverless cars.(Business). *The Times (London, England)* (2015).

[16] Sachin Farfade, Mohammad J. Saberian, and Li-Jia Li. 2015. Multi-view Face Detection Using Deep Convolutional Neural Networks. (02 2015). https://doi.org/10.1145/2671188.2749408

[17] Cong Geng and Xudong Jiang. 2009. Face Recognition Using SIFT Features. In *Proceedings of the 16th IEEE International Conference on Image Processing (ICIP'09)*. IEEE Press, Piscataway, NJ, USA, 3277–3280. http://dl.acm.org/citation.cfm?id=1819298.1819645

[18] E. U. Haq, X. Huarong, and M. I. Khattak. 2017. Face Recognition by SVM Using Local Binary Patterns. In *2017 14th Web Information Systems and Applications Conference (WISA)*. 172–175. https://doi.org/10.1109/WISA.2017.68

[19] Kaiming He, Xiangyu Zhang, Shaoqing Ren, and Jian Sun. 2015. Deep Residual Learning for Image Recognition. *CoRR* abs/1512.03385 (2015). arXiv:1512.03385 http://arxiv.org/abs/1512.03385 [20] Balazs Holczer. [n. d.]. ([n. d.]).

[21] Andrew G. Howard, Menglong Zhu, Bo Chen, Dmitry Kalenichenko, Weijun Wang, Tobias Weyand, Marco Andreetto, and Hartwig Adam. 2017. MobileNets: Efficient Convolutional Neural Networks for Mobile Vision Applications. *CoRR* abs/1704.04861 (2017). arXiv:1704.04861 http://arxiv.org/abs/1704.04861

[22] Peiyun Hu and Deva Ramanan. 2016. Finding Tiny Faces. *CoRR* abs/1612.04402 (2016). arXiv:1612.04402 http://arxiv.org/abs/1612.04402

[23] Amin Jalali, Rammohan Mallipeddi, and Minho Lee. 2017. Sensitive Deep Convolutional Neural Network for Face Recognition at Large Standoffs with Small Dataset. *Expert Syst. Appl.* 87, C (Nov. 2017), 304–315. https://doi.org/10.1016/j.eswa.2017.06.025

[24] Bart Jansen. 2018. Orlando airport unveils face scans.(LIFE)(Orlando International Airport)(Brief article). *USA Today* (2018).
[25] Bong-Nam Kang, Yonghyun Kim, and D. Kim. 2016. Deep convolution neural network with stacks of multi-scale convolutional layer block using triplet of faces for face recognition in the wild. In *2016 IEEE International Conference on Systems, Man, and Cybernetics (SMC)*. 004460–004465. https://doi.org/10.1109/SMC.2016.7844934
[26] Ira Kemelmacher-Shlizerman, Steven M. Seitz, Daniel Miller, and Evan Brossard. 2015. The MegaFace Benchmark: 1 Million Faces for Recognition at Scale. *CoRR* abs/1512.00596 (2015). arXiv:1512.00596 http://arxiv.org/abs/1512.00596 [27] B. Kitchenham and S Charters. 2007. Guidelines for performing Systematic Literature Reviews in Software Engineering.
[28] Alex Krizhevsky, Ilya Sutskever, and Geoffrey E. Hinton. 2012. ImageNet Classification with Deep Convolutional Neural Networks. In *Proceedings of the 25th International Conference on Neural Information Processing Systems - Volume 1 (NIPS'12)*. Curran Associates Inc., USA, 1097–1105. http://dl.acm.org/citation.cfm?id=2999134.2999257
[29] Y. Li, Z. Mu, and T. Zhang. 2016. Local feature extraction and recognition under expression variations based on multimodal face and ear spherical map. In *2016 9th International Congress on Image and Signal Processing, BioMedical Engineering and Informatics (CISP-BMEI)*. 286–290. https://doi.org/10.1109/CISP-BMEI.2016.7852723
[30] Tsung-Yi Lin, Priya Goyal, Ross B. Girshick, Kaiming He, and Piotr Dollár. 2017. Focal Loss for Dense Object Detection. *CoRR* abs/1708.02002 (2017). arXiv:1708.02002 http://arxiv.org/abs/1708.02002
[31] Mamta and Madasu Hanmandlu. 2014. Robust authentication using the unconstrained infrared face images. *Expert Systems with Applications* 41, 14 (2014), 6494 – 6511. https://doi.org/10.1016/j.eswa.2014.03.040
[32] I. Masi, F. Chang, J. Choi, S. Harel, J. Kim, K. Kim, J. Leksut, S. Rawls, Y. Wu, T. Hassner, W. AbdAlmageed, G. Medioni, L. Morency, P. Natarajan, and R. Nevatia. 2018. Learning Pose-Aware Models for Pose-Invariant Face Recognition in the Wild. *IEEE Transactions on Pattern Analysis and Machine Intelligence* (2018), 1–1. https://doi.org/10.1109/TPAMI. 2018.2792452
[33] T. M. Murphy, R. Broussard, R. Schultz, R. Rakvic, and H. Ngo. 2017. Face detection with a ViolaâĂŞJones based hybrid network. *IET Biometrics* 6, 3 (2017), 200–210. https://doi.org/10.1049/iet-bmt.2016.0037
[34] Mahyar Najibi, Pouya Samangouei, Rama Chellappa, and Larry S. Davis. 2017. SSH: Single Stage Headless Face Detector. *CoRR* abs/1708.03979 (2017). arXiv:1708.03979 http://arxiv.org/abs/1708.03979
[35] Andrew Ng. 2015. What data scientists should know about deep learning. (2015). https://www.slideshare.net/ExtractConf.
[36] S. Pannirselvam and S. Prasath. 2015. A Novel Technique for Face Recognition and Retrieval Using Fiducial Point Features. *Procedia Computer Science* 47 (2015), 301 – 310. https://doi.org/10.1016/j.procs.2015.03.210 Graph Algorithms, High Performance Implementations and Its Applications ( ICGHIA 2014 ).
[37] P. J. Phillips, A. N. Yates, J. R. Beveridge, and G. Givens. 2017. Predicting Face Recognition Performance in Unconstrained Environments. In *2017 IEEE Conference on Computer Vision and Pattern Recognition Workshops (CVPRW)*. 557–565. https://doi.org/10.1109/CVPRW.2017.83
[38] R. Ranjan, S. Sankaranarayanan, A. Bansal, N. Bodla, J. Chen, V. M. Patel, C. D. Castillo, and R. Chellappa. 2018. Deep Learning for Understanding Faces: Machines May Be Just as Good, or Better, than Humans. *IEEE Signal Processing Magazine* 35, 1 (Jan 2018), 66–83. https://doi.org/10.1109/MSP.2017.2764116
[39] Yasar Rehman, Po Lai Man, and Mengyang Liu. 2018. LiveNet: Improving Features Generalization for Face Liveness Detection using Convolution Neural Networks. *Expert Systems with Applications* 108 (05 2018). https://doi.org/10.1016/j.eswa.2018.05.004
[40] Shaoqing Ren, Kaiming He, Ross B. Girshick, and Jian Sun. 2015. Faster R-CNN: Towards Real-Time Object Detection with Region Proposal Networks. *CoRR* abs/1506.01497 (2015). arXiv:1506.01497 http://arxiv.org/abs/1506.01497 [41] Timothy Revell. 2016. Chinese town uses your face as a ticket. *New Scientist* 232, 3101 (2016), 25–25.
[42] David Robertson, Robin Kramer, and A Burton. 2015. Face Averages Enhance User Recognition for Smartphone Security. *PLoS One* 10, 3 (2015). http://search.proquest.com/docview/1666740493/
[43] Adrian Rosebrock. 2017. Deep Learning for Computer Vision With Python. 1 (2017).
[44] Florian Schroff, Dmitry Kalenichenko, and James Philbin. 2015. FaceNet: A Unified Embedding for Face Recognition and Clustering. *CoRR* abs/1503.03832 (2015). arXiv:1503.03832 http://arxiv.org/abs/1503.03832
[45] Florian Schroff, Dmitry Kalenichenko, and James Philbin. 2015. FaceNet: A unified embedding for face recognition and clustering. (06 2015), 815–823. https://doi.org/10.1109/CVPR.2015.7298682

[46] Radhey Shyam and Yogendra Narain Singh. 2016. Recognizing Individuals from Unconstrained Facial Images. In *Intelligent Systems Technologies and Applications*, Stefano Berretti, Sabu M. Thampi, and Praveen Ranjan Srivastava (Eds.). Springer International Publishing, Cham, 383–392.

[47] Karen Simonyan and Andrew Zisserman. 2014. Very Deep Convolutional Networks for Large-Scale Image Recognition. *CoRR* abs/1409.1556 (2014). arXiv:1409.1556 http://arxiv.org/abs/1409.1556

[48] M. Singh, S. Nagpal, N. Gupta, S. Gupta, S. Ghosh, R. Singh, and M. Vatsa. 2016. Cross-spectral cross-resolution video database for face recognition. In *2016 IEEE 8th International Conference on Biometrics Theory, Applications and Systems (BTAS)*. 1–7. https://doi.org/10.1109/BTAS.2016.7791166

[49] Ya Su. 2018. Robust Video Face Recognition Under Pose Variation. *Neural Process. Lett.* 47, 1 (Feb. 2018), 277–291. https://doi.org/10.1007/s11063-017-9649-8

[50] Ya Su. 2018. Robust Video Face Recognition Under Pose Variation. *Neural Processing Letters* 47, 1 (01 Feb 2018), 277–291. https://doi.org/10.1007/s11063-017-9649-8

[51] Yi Sun, Xiaogang Wang, and Xiaoou Tang. 2014. Deeply learned face representations are sparse, selective, and robust. *CoRR* abs/1412.1265 (2014). arXiv:1412.1265 http://arxiv.org/abs/1412.1265

[52] Christian Szegedy, Wei Liu, Yangqing Jia, Pierre Sermanet, Scott E. Reed, Dragomir Anguelov, Dumitru Erhan, Vincent Vanhoucke, and Andrew Rabinovich. 2014. Going Deeper with Convolutions. *CoRR* abs/1409.4842 (2014). arXiv:1409.4842 http://arxiv.org/abs/1409.4842

[53] Christian Szegedy, Wei Liu, Yangqing Jia, Pierre Sermanet, Scott E. Reed, Dragomir Anguelov, Dumitru Erhan, Vincent Vanhoucke, and Andrew Rabinovich. 2014. Going Deeper with Convolutions. *CoRR* abs/1409.4842 (2014). arXiv:1409.4842 http://arxiv.org/abs/1409.4842

[54] Y. Taigman, M. Yang, M. Ranzato, and L. Wolf. 2014. DeepFace: Closing the Gap to Human-Level Performance in Face Verification. In *2014 IEEE Conference on Computer Vision and Pattern Recognition*. 1701–1708. https://doi.org/10.1109/ CVPR.2014.220

[55] Paul Viola and Michael Jones. 2001. Rapid Object Detection using a Boosted Cascade of Simple Features. 1 (02 2001), I–511.

[56] W. Wan and J. Chen. 2017. Occlusion robust face recognition based on mask learning. In *2017 IEEE International Conference on Image Processing (ICIP)*. 3795–3799. https://doi.org/10.1109/ICIP.2017.8296992

[57] Hao Wang, Yitong Wang, Zheng Zhou, Xing Ji, Zhifeng Li, Dihong Gong, Jingchao Zhou, and Wei Liu. 2018. CosFace: Large Margin Cosine Loss for Deep Face Recognition. *CoRR* abs/1801.09414 (2018). arXiv:1801.09414 http://arxiv.org/abs/1801.09414

[58] Jianfeng Wang, Ye Yuan, and Gang Yu. 2017. Face Attention Network: An Effective Face Detector for the Occluded Faces. *CoRR* abs/1711.07246 (2017). arXiv:1711.07246 http://arxiv.org/abs/1711.07246

[59] Michael J. Wilber, Vitaly Shmatikov, and Serge J. Belongie. 2016. Can we still avoid automatic face detection? *CoRR* abs/1602.04504 (2016). arXiv:1602.04504 http://arxiv.org/abs/1602.04504

[60] Claes Wohlin. 2014. Guidelines for Snowballing in Systematic Literature Studies and a Replication in Software Engineering. In *Proceedings of the 18th International Conference on Evaluation and Assessment in Software Engineering (EASE '14)*. ACM, New York, NY, USA, Article 38, 10 pages. https://doi.org/10.1145/2601248.2601268

[61] S. Yang, P. Luo, C. C. Loy, and X. Tang. 2016. WIDER FACE: A Face Detection Benchmark. In *2016 IEEE Conference on Computer Vision and Pattern Recognition (CVPR)*. 5525–5533. https://doi.org/10.1109/CVPR.2016.596

[62] Shuo Yang, Yuanjun Xiong, Chen Change Loy, and Xiaoou Tang. 2017. Face Detection through Scale-Friendly Deep Convolutional Networks. *CoRR* abs/1706.02863 (2017). arXiv:1706.02863 http://arxiv.org/abs/1706.02863

[63] Nanhai Zhang and Weihong Deng. 2016. Fine-grained LFW database. In *2016 International Conference on Biometrics (ICB)*. 1–6. https://doi.org/10.1109/ICB.2016.7550057

[64] Shifeng Zhang, Xiangyu Zhu, Zhen Lei, Hailin Shi, Xiaobo Wang, and Stan Z. Li. 2017. $S^3$FD: Single Shot Scale-invariant Face Detector. *CoRR* abs/1708.05237 (2017). arXiv:1708.05237 http://arxiv.org/abs/1708.05237

[65] Tianyue Zheng, Weihong Deng, and Jiani Hu. 2017. Cross-Age LFW: A Database for Studying Cross-Age Face Recognition in Unconstrained Environments. *CoRR* abs/1708.08197 (2017). arXiv:1708.08197 http://arxiv.org/abs/1708. 08197

[66] Tianyue Zheng, Weihong Deng, and Jiani Hu. 2017. Cross-Age LFW: A Database for Studying Cross-Age Face Recognition in Unconstrained Environments. *CoRR* abs/1708.08197 (2017). arXiv:1708.08197 http://arxiv.org/abs/1708. 08197